%% file: main.tex
\documentclass[11pt,a4paper]{article}
\usepackage{acl}
\usepackage{times,latexsym}
\usepackage{url}
\usepackage[T1]{fontenc}
\usepackage{bm}
\usepackage{amsmath,amssymb,amsfonts}
\usepackage{multicol}
\usepackage{multirow}
\usepackage{float}
\usepackage{booktabs}
\usepackage{graphics}
\usepackage{xcolor}

\newcommand{\stitle}[1]{\smallskip\noindent{\bf #1}}

\title{Continual Contrastive Finetuning Improves\\ Low-Resource Relation Extraction}

\author{
Wenxuan Zhou$^{\dagger}$, Sheng Zhang$^{\ddagger}$, Tristan Naumann$^{\ddagger}$, Muhao Chen$^{\dagger}$ \and Hoifung Poon
$^{\ddagger}$\\
$^{\dagger}$University of Southern California, $^{\ddagger}$Microsoft Research \\ 
\texttt{\{zhouwenx,muhaoche\}@usc.edu}\\
\texttt{\{zhang.sheng,tristan,hoifung\}@microsoft.com}\\
}

\date{}

\begin{document}
\maketitle
\input{introduction}
\input{related_work}

\input{method}
\input{experiments}
\input{conclusion}

\bibliography{tacl2021}
\input{appendix}

\end{document}

%% file: introduction.tex
\begin{abstract}
Relation extraction (RE), which has relied on structurally annotated corpora for model training, has been particularly challenging in low-resource scenarios and domains.
Recent literature has tackled low-resource RE by self-supervised learning, where the solution involves pretraining the entity pair embedding by RE-based objective and finetuning on labeled data by classification-based objective.
However, a critical challenge to this approach is the gap in objectives, which prevents the RE model from fully utilizing the knowledge in pretrained representations.
In this paper, we aim at bridging the gap and propose to pretrain and finetune the RE model using consistent objectives of contrastive learning.
Since in this kind of representation learning paradigm, one relation may easily form multiple clusters in the representation space, we further propose a multi-center contrastive loss that allows one relation to form multiple clusters to better align with pretraining.
Experiments on two document-level RE datasets, BioRED and Re-DocRED, demonstrate the effectiveness of our method.
Particularly, when using 1\% end-task training data, our method outperforms PLM-based RE classifier by 10.5\% and 6.1\% on the two datasets, respectively.
\end{abstract}

\section{Introduction}
Relation extraction (RE) is a fundamental task in NLP.
It aims to identify the relations among entities in a given text from a predefined set of relations.
While much effort has been devoted to RE in supervised settings~\cite{zhang-etal-2017-position,zhang-etal-2018-graph,nan-etal-2020-reasoning}, RE is extremely challenging in high-stakes domains such as biology and medicine, where 
annotated data are comparatively scarce due to overly high annotation costs.
Therefore, there is a practical and urgent need for developing low-resource RE models 
without the reliance on large-scale end-task annotations.

To realize low-resource RE, previous work has focused on pretraining entity pair embedding on large corpora using RE-based pretraining objectives.
Particularly, \citet{baldini-soares-etal-2019-matching} propose a self-supervised matching-the-blanks (MTB) objective that encourages embeddings of the same entity pairs in different sentences to be similar.
Later work~\cite{peng-etal-2020-learning,qin-etal-2021-erica} extends this idea with distant supervision~\cite{mintz-etal-2009-distant} and improves representation learning using contrastive learning~\cite{hadsell2006dimensionality,oord2018representation,chen2020simple}.
To adapt to 
training on RE annotations, these works finetune pretrained entity pair embedding on labeled data using classification-based objectives.
Although this paradigm produces better results compared to RE models initialized with pretrained language models~(PLMs), it creates a significant divergence between pretraining and finetuning objectives, 
thus preventing the model from fully exploiting knowledge in pretraining.

In this paper, we aim to bridge this gap in RE pretraining and finetuning.
Our key idea is to use similar objectives in pretraining and finetuning.
First, we propose to continually finetune pretrained embedding by contrastive learning, which encourages the entity pair embeddings corresponding to the same relation to be similar.
However, as pretraining and finetuning are conducted on different tasks, entity pairs of the same relation can form multiple different clusters in the pretrained embedding, where standard supervised contrastive loss~\cite{khosla2020supervised} may distort the representation because of its underlying one-cluster assumption~\cite{graf2021dissecting}.
Therefore, we further propose a multi-center contrastive loss~(MCCL), which encourages an entity pair to be similar to only a subset of entity pairs of the same relation, allowing one relation to form multiple clusters.
Second, we propose to use classwise k-nearest neighbors~(kNN; \citealt{khandelwal2019generalization,khandelwal2020nearest}) in inference, where predictions are made based on most similar instances.

We focus our work on document-level RE~\cite{jia-etal-2019-document,yao-etal-2019-docred}, which consists of both intra- and cross-sentence relations. 
To the best of our knowledge, this work represents the first effort to explore self-supervised pretraining for document-level RE.
Unlike prior studies~\cite{peng-etal-2020-learning,qin-etal-2021-erica}, we do not use distant supervision.
Instead, we pretrain entity pair embedding with an improved MTB objective on unlabeled corpora, where we use contrastive learning to learn representations that suit downstream RE. 
We then finetune the pretrained model on labeled data with MCCL.
Experiments on two datasets, BioRED~\cite{luo2022biored} in the biomedical domain and Re-DocRED~\cite{tan2022revisiting} in the general domain, demonstrate that our pretraining and finetuning objectives significantly outperform baseline methods in low-resource settings.
Particularly, in the low-resource setting of using 1\% of labeled data, our method outperforms PLM-based classifiers by 10.5\% and 6.1\% on BioRED and Re-DocRED, respectively.
Based on our pretrained representations, MCCL outperforms classification-based finetuning by 6.0\% and 4.1\%, respectively.
We also find observe that as more data becomes available, the performance gap between MCCL and classification-based finetuning diminishes.

Our technical contributions are three-fold. First, we propose to pretrain the PLMs based on our improved MTB objective and show that it significantly improves PLM performance in low-resource document-level RE. Second, we present a technique that bridges the gap of learning objectives between RE pretraining and finetuning with continual contrastive finetuning and kNN-based inference, helping the RE model leverage pretraining knowledge. Third, we design a novel MCCL finetuning objective, allowing one relation to form multiple different clusters, thus further reducing the distributional gap between pretraining and finetuning.

%% file: related_work.tex
\section{Related Work}


\stitle{Document-level RE.}
Existing document-level RE models can be classified into graph-based and sequence-based models.
Graph-based models construct document graphs spanning across sentence boundaries and use graph encoders such as the graph convolution network~(GCN; \citealt{kipf2017semi}) to aggregate information.
Particularly, \citet{quirk-poon-2017-distant} build document graphs using words as nodes with inner- and inter-sentence dependencies~(e.g., syntactic dependencies, coreference, etc.) as edges.
Later work extends this idea by applying different network structures~\cite{peng-etal-2017-cross,jia-etal-2019-document} or introducing other node types and edges~\cite{christopoulou-etal-2019-connecting,nan-etal-2020-reasoning,zeng-etal-2020-double}.
On the other hand, sequence-based methods~\cite{zhou2021document,Zhang2021DocumentlevelRE,tan-etal-2022-document} use PLMs to learn cross-sentence dependencies without using graph structures.
Particularly, \citet{zhou2021document} propose to enrich relation mention representation by localized context pooling.
\citet{Zhang2021DocumentlevelRE} propose to model the inter-dependencies between relation mentions by semantic segmentation~\cite{ronneberger2015u}.
In this work, we study a general method of self-supervised RE.
Therefore, our method is independent of the model architecture and can be adapted to different RE models.

\stitle{Low-resource RE.}
Labeled RE data may be scarce in real-world applications, especially 
in low-resource and high-stakes domains such as finance and biomedicine.
Much effort has been devoted to training RE models in low-resource settings.
Some work tackles low-resource RE by indirect supervision, which solves RE by other tasks such as machine reading comprehension~\cite{levy-etal-2017-zero}, textual entailment \cite{sainz-etal-2021-label}, and abstractive summarization \cite{lu2022summarization}.
However, indirect supervision may not be practical in high-stake domains, where annotated data for other tasks are also scarce.
Other efforts~\cite{baldini-soares-etal-2019-matching,peng-etal-2020-learning,qin-etal-2021-erica} improve low-resource RE by pretraining on large corpora with RE-based objectives.
Specifically, \citet{baldini-soares-etal-2019-matching} propose an MTB objective that encourages embeddings of the same entity pairs in different sentences to be similar.
\citet{peng-etal-2020-learning} propose to pretrain on distantly labeled corpora, where they make embeddings of entity pairs with the same distant label to be similar.
They also introduce a contrastive learning based training objective to improve representation learning.
\citet{qin-etal-2021-erica} further introduce an entity discrimination task and pretrain the RE model on distantly labeled document corpora.
In this paper, we study self-supervised pretraining for document-level RE.
We study how to reduce the gap between pretraining and finetuning, which is critical to bridge the training signals obtained in these two stages but has been overlooked in prior work.

%% file: method.tex
\section{Method}

\input{problem_definition}

Our training pipeline consists of two phases: pretraining and finetuning.
In pretraining, we use the (unlabeled) document corpus to pretrain the entity pair embedding based on our improved \textit{matching-the-blanks} training objective~(MTB; \citealt{baldini-soares-etal-2019-matching}), where the LM learns to decide whether two entity pair embeddings correspond to the entity pairs or not,
and the learning of representation is enhanced with contrastive learning.
In finetuning, we continue to train the pretrained model on relation-labeled data using a multi-center contrastive loss (MCCL), which 
achieves better performance than the traditional classifier paradigm
due to its better-aligned learning objective with pretraining.
After training, we use classwise k-nearest neighbor~(kNN) inference that suits well the contrastively finetuned model.

The rest of this section is organized as follows: we introduce the model architecture used in both pretraining and finetuning in Section~\ref{ssec:model_architecture}, the pretraining process in Section~\ref{ssec:pretraining}, finetuning in Section~\ref{ssec:finetuning}, and inference in Section~\ref{ssec:inference}.

\input{methods/method-architecture}
\input{methods/method-pretrain}
\input{methods/method-finetune}
\input{methods/method-inference}

%% file: problem_definition.tex
In this work, we study a self-supervised approach for document-level RE.
Given a document $d$ and a set of entities $\{e_i\}_{i=1}^N$, where each entity $e_i$ has one or multiple entity mentions in the document, document-level RE aims at predicting the relations of all entity pairs $(e_s, e_o)_{s, o \,\in\, \{1, ..., N\}}$ from a predefined set of relationships $\mathcal{R}$~(including an \textsc{na} class indicating no relation exists), where $e_s$ and $e_o$ are the subject and object entities, respectively.
In the self-supervised RE setting, we have a large unlabeled document corpus for pretraining and a labeled RE dataset for finetuning.
The document corpus has been annotated with entity mentions and the associated entity types but no relations.
Our goal is to train a document-level RE classifier, especially in the low-resource setting.

%% file: methods/method-architecture.tex
\subsection{Model Architecture}
\label{ssec:model_architecture}
\stitle{Encoder.} Given a document $d=[x_1, x_2, ..., x_l]$, we first mark the spans of the entity mentions by adding special entity markers \textsc{[E]} and \textsc{[/E]} to the start and the end of each mention.
Then we encode the document with a PLM to get the contextual embedding of textual tokens:
\begin{equation*}
    \bm{H} = [\bm{h}_1, \bm{h}_2, ..., \bm{h}_l]=\text{PLM}\left([x_1, x_2, ..., x_l]\right).
\end{equation*}
We take the contextual embedding of \textsc{[E]} at the last layer of the PLM as the embedding of entity mentions.
We accumulate the embedding of mentions corresponding to the same entity by LogSumExp pooling~\cite{jia-etal-2019-document} to get the entity embedding $\bm{h}_{e_i}$.

\stitle{Entity pair embedding.}
Given an entity pair $t=(e_s, e_o)$ in document $d$, where $e_s$ and $e_o$ are 
the subject and object entities, respectively, we calculate the entity pair embedding by:
\begin{equation*}
    \bm{z}^t=\bm{W}_\text{linear}\left[\bm{h}_{e_s}, \bm{h}_{e_o}, \bm{c}^{(e_s, e_o)}\right].
\end{equation*}
Here $\bm{h}_{e_s}, \bm{h}_{e_o}\in \mathbb{R}^d$ are embeddings of subject and object entities, $\bm{c}_{e_s, e_o}\in \mathbb{R}^d$ is the localized context encoding for $(e_s, e_o)$, $\bm{W}_\text{linear} \in \mathbb{R}^{3d\times d}$ is a linear projector.
The localized context encoding is introduced by~\citet{zhou2021document} to derive 
the context embedding conditioned on an entity pair,
which finds the context that both the subject and object entities attend to.
Specifically, denote the multi-head attention in the last layer of PLM as $\bm{A}\in\mathbb{R}^{m \times l \times l}$, where $m$ is the number of attention heads, $l$ is the input length, we first take the attention scores from \textsc{[E]} as the attention from each entity mention, then accumulate the attention of this entity mention 
by mean pooling to get the entity-level attention $\bm{A}^{(e_i)}\in \mathbb{R}^{m \times l}$. Finally, we compute $\bm{c}^{(e_s,e_o)}$ by:
\begin{align*}
    \bm{A}^{(e_s,e_o)} &= \bm{A}^{(e_s)} \odot \bm{A}^{(e_o)}, \\
    \bm{q}^{(e_s,e_o)} &= \sum_{i=1}^m \bm{A}^{(e_s,e_o)}_i, \\
    \bm{a}^{(e_s,e_o)} &= \bm{q}^{(e_s,e_o)} / {\bm{1}^\intercal \bm{q}^{(e_s,e_o)}}, \\
    \bm{c}^{(e_s,e_o)} &= \bm{H}^\intercal \bm{a}^{(e_s,e_o)}.
\end{align*}
We introduce in the rest of the section how to pretrain and finetune the RE model based on the entity pair embedding $\bm{z}^{(e_s, e_o)}$.

%% file: methods/method-pretrain.tex
\subsection{Pretraining}
\label{ssec:pretraining}
We pretrain the LM on the document corpus using the MTB objective.
MTB is based on a simple assumption that, 
in contrast to different entity pairs,
it is more frequent for the same entity pair to be connected with the same relation. 
The MTB objective transforms the similarity learning problem into a pairwise binary classification problem: given two relation-describing utterances where entity mentions are masked, the model classifies whether the entity pairs are the same or not.
This pretraining objective has shown effectiveness in several sentence-level RE datasets\cite{zhang-etal-2017-position,hendrickx-etal-2010-semeval,han-etal-2018-fewrel}.

However, when it comes to document-level RE, \citet{qin-etal-2021-erica} have observed no improvement
led by the vanilla MTB pretraining. 
Therefore, we replace the pairwise binary classification with contrastive learning, which is adopted in later RE pretraining works~\cite{peng-etal-2020-learning,qin-etal-2021-erica} and can effectively learn from more positive and negative examples.
Details of training objectives are elaborated in the rest of the section.
We introduce the details of data preprocessing of the pretraining corpus in Appendix~\ref{ssec:data_preparation}.

\stitle{Training objective.}
The overall goal of pretraining is to make the embedding of the same entity pair from different documents more similar than different entity pairs.
For clarity, we call two same entity pairs from different documents as a positive pair, and two different entity pairs as a negative pair.
We use the InfoNCE loss~\cite{oord2018representation} to model this objective.
Given the documents in batch, $\mathcal{P}$ as the set of all positive pairs, and $\mathcal{N}_t$ denote the set of entity pairs different to $t$, the contrastive MTB loss is\footnote{Similar to~\citet{baldini-soares-etal-2019-matching}, we randomly mask the entities in documents with a probability of 0.7 to avoid shortcut learning.}:
\begin{align}
    \mathcal{L}_\text{rel}&=-\frac{1}{|\mathcal{P}|} \sum_{t_i, t_j\in \mathcal{P}} \log \frac{e^{\text{sim} (\bm{z}^{t_i}, \bm{z}^{t_j})/\tau}}{\mathcal{Z}_{t_i}}, \label{eq:contra}\\
    \mathcal{Z}_{t_i} &= e^{\text{sim} (\bm{z}^{t_i}, \bm{z}^{t_j})/\tau} + \sum_{t_k \in \mathcal{N}_{t_i}} e^{\text{sim} (\bm{z}^{t_i}, \bm{z}^{t_k})/\tau} \nonumber,
\end{align}
where $\text{sim}(\bm{z}^{t_i}, \bm{z}^{t_j})$ denotes the similarity between the embeddings of $t_i$ and $t_j$, and $\tau$ is a temperature hyperprameter.
Following~\citet{chen2020simple}, we use cosine similarity as the similarity metric.
Similar to SimCSE~\cite{gao-etal-2021-simcse}, we further add a self-supervised contrastive loss that requires the same entity pair embedding augmented by different dropout masks to be similar, thus encouraging the model to learn more instance-discriminative features that lead to less collapsed representations.
Specifically, denote the two entity pair embeddings of $t$ derived by different dropout masks as $\bm{z}^t$ and $\hat{\bm{z}}^t$, respectively, the set of all entity pairs in the batch as $\mathcal{T}$, and the set of entity pairs in positive pairs as $\mathcal{T}_P$, the self-supervised loss is:
\begin{align*}
    \mathcal{L}_\text{self}&=-\frac{1}{|\mathcal{T}_P|} \sum_{t_i \in \mathcal{T}_P} \log \frac{e^{\text{sim} (\bm{z}^{t_i}, \hat{\bm{z}}^{t_i})/\tau}}{\mathcal{Z}_{t_i}}, \\
    \mathcal{Z}_{t_i} &= e^{\text{sim} (\bm{z}^{t_i}, \hat{\bm{z}}^{t_i})/\tau} +\sum_{t_k \in \mathcal{T} \backslash \{t_i\}} e^{\text{sim}(\bm{z}^{t_i}, \hat{\bm{z}}^{t_k})/\tau}.
\end{align*}
Finally, we use a masked language model loss $\mathcal{L}_\text{mlm}$ to adapt the LM to the document corpus.
The overall pretraining objective is:
\begin{equation*}
    \mathcal{L}_\text{pretrain}=\mathcal{L}_\text{rel} + \mathcal{L}_\text{self} + \mathcal{L}_\text{mlm}.
\end{equation*}
For faster convergence, we initialize our model with a PLM that is pretrained on a larger corpus, and continually pretrain the PLM on the document corpus with our new pretraining objectives.
We use BERT~\cite{devlin-etal-2019-bert} for the general domain and PubmedBERT~\cite{pubmedbert} for the biomedical domain.

%% file: methods/method-finetune.tex
\subsection{Finetuning}
\label{ssec:finetuning}
After pretraining, we finetune the LM on labeled document-level RE datasets.
In previous studies~\cite{baldini-soares-etal-2019-matching,peng-etal-2020-learning,qin-etal-2021-erica}, pretraining and finetuning are conducted in processes with different learning objectives. Specifically, after using the pretrained weights to initialize a RE classifier, the model is finetuned with a classification-based training objective.
Based on our model architecture, a straightforward finetuning method is to add a 
softmax classifier upon the entity pair embedding, for which a cross-entropy loss for a batch of entity pairs $\mathcal{T}$ is formulated as:
\begin{align*}
    P_r^{t_i}&=\text{softmax}(\bm{W}_r \bm{z}^{t_i} + b_r), \\
    \mathcal{L}_\text{ce} &= - \frac{1}{|\mathcal{T}|}\sum_{t_i\in  \mathcal{T}}\log(P_{y_{t_i}}^{t_i}),
\end{align*}
where $y_t$ is the ground-truth label for entity pair $t$, $\bm{W}_r, b_r$ are the weight and bias of the classifier.
Though this approach has shown improvements, it may produce sub-optimal outcomes from MTB pretraining since it implicitly assumes that entity pairs corresponding to the same relation are in the same cluster, while MTB pretraining may learn multiple clusters for a relation.
For example, the entity pairs \emph{(Honda Corp., Japan)} and \emph{(Mount Fuji, Japan)}, 
although likely to be expressed with the same relation \emph{country} in documents, are likely to be in different clusters since MTB views them as negative pairs due to different subject entities.
Therefore, we propose an MCCL objective that can bridge these gaps.
Next, we will discuss the distributional assumption of the softmax classifier as well as supervised contrastive loss, then present our MCCL objective.

\begin{table}[!t]
    \centering
    \scalebox{0.87}{
    \begin{tabular}{lccc}
    \toprule
     Classifier & BioRED& Re-DocRED\\
     \midrule
     \multicolumn{3}{c}{\emph{One-cluster}} \\
     Softmax& 28.6& 39.3\\
     Nearest centroid& 12.5& 4.1\\
     \midrule
     \multicolumn{3}{c}{\emph{Multi-cluster}} \\
     classwise kNN & \textbf{36.7}& \textbf{54.1}\\
    \bottomrule
    \end{tabular}}
    \caption{Probing results (in $F_1$) on the test set of BioRED and Re-DocRED.}
    \label{tab:probing}
\end{table}

\stitle{Distributional assumption.}
We conduct a probing analysis on the distribution of pretrained representations to further justify the multi-cluster assumption.
Specifically, we fix the weights of the pretrained MTB model and fit different classifiers on top of it, including a softmax classifier, a nearest centroid classifier (both assuming one cluster for a relation), and a classwise kNN classifier (assuming multiple clusters for a relation).
We evaluate these classifiers on the test set.
Results are shown in Table~\ref{tab:probing}.
We find that classwise kNN greatly outperforms others, showing that MTB pretraining learns multiple clusters for a relation.

Therefore, to accommodate this multi-cluster assumption, we need to finetune the representations with a training objective that suits multiple clusters for each relation.
Beside using the softmax classifier with cross-entropy loss, we also consider supervised contrastive loss (SupCon; \citealt{khosla2020supervised,gunel2020supervised}).
SupCon has a similar loss form to InfoNCE in Eq.~\eqref{eq:contra}, except that it uses instances of the same/different relations as positive/negative pairs.
However, previous work~\cite{graf2021dissecting} has shown that both softmax and SupCon are minimized when the representations of each class collapse to the vertex of a regular simplex.
In our case, this means the entity pair embeddings corresponding to the same relation in pretraining collapses to a single point, which creates a distributional gap between pretraining and finetuning.

\stitle{Training objective.}
We thereby propose the MCCL objective.
Given entity pairs $\mathcal{T}$ and sets of entity pairs grouped by their relations $\{{\mathcal{T}_r}\}_{r\in \mathcal{R}}$, our loss is formulated as:
\begin{align*}
    w_r^{(t_i, t_j)} &= \frac{e^{\text{sim}(\bm{z}^{t_i}, \bm{z}^{t_j}) / \tau_1}}{\sum_{t_k \in \mathcal{T}_r \backslash \{t_i\}}e^{\text{sim}(\bm{z}^{t_i}, \bm{z}^{t_k})/\tau_1}}, \\
    s_r^{t_i} &= \sum_{t_j \in \mathcal{T}_r \backslash \{t_i\}} w_r^{(t_i, t_j)} {\text{sim}(\bm{z}^{t_i}, \bm{z}^{t_j})}, \\
    P_r^{t_i}&=\text{softmax}((s_r^{t_i} + b_r) / \tau_2), \\
    \mathcal{L}_\text{mccl} &= -\frac{1}{|\mathcal{T}|}\sum_{t_i\in  \mathcal{T}}\log(P_{y_{t_i}}^{t_i}),
\end{align*}
where $\tau_1$ and $\tau_2$ are temperature hyperparameters, $b_r\in \mathbb{R}$ is the classwise bias.
The loss calculation can be split into two steps.
First, we calculate the similarity between $t_i$ and relation $r$, which is a weighted average of the similarity between $t_i$ and $t_j\in \mathcal{T}_r$ such that a more similar $t_j$ has a larger weight.
Next, we use the cross-entropy loss to make the similarity of ground-truth relation larger than others.
In this way, MCCL only optimizes $t_i$ to be similar to a few closest entity pairs of the ground-truth relation, and thus encourages multiple clusters in entity pair embedding.
Note that MCCL can be easily extended to support multi-label classification scenarios, for which details are given in Appendix~\ref{ssec:multilabel_RE}.

\stitle{Proxies.}
We use batched training for finetuning, where entity pairs in the current batch are used to calculate MCCL.
However, it is possible that a subset of relations in $\mathcal{R}$, especially the long-tail relations, are rare 
or missing in the current batch.
When $\mathcal{T}_r \backslash \{t_i\}$ is empty, $s_r^{t_i}$ and MCCL become undefined.
To tackle this problem, we propose the use of proxies~\cite{movshovitz2017no,zhu2022balanced}.
We add one proxy vector $p_r$ for each relation $r$, which is a trainable parameter and associated with an embedding $\bm{z}_r^p$.
We incorporate the proxies into MCCL by changing $\mathcal{T}_r$ to $\mathcal{T}_r'=\mathcal{T}_r \cup \{p_r\}$, ensuring that $\mathcal{T}_r' \backslash \{t_i\}$ is never empty in training and preventing MCCL from becoming undefined.
The proxies are randomly initialized and updated during training by backward propagation.

%% file: methods/method-inference.tex
\subsection{Inference}
\label{ssec:inference}
We use the classwise kNN~\cite{christobel2013new} for inference,
which predicts relations based on similarly represented instances and thus aligns with our contrastive finetuning objective.
Given a new entity pair to predict, we first find $k$ most similar instances\footnote{Measured by cosine similarity. If a relation has fewer than $k$ entity pairs in training data, we use all of them.} in the training data of each relation (including \textsc{na}), then calculate the average cosine similarity of each relation $s^\text{avg}_r$. 
Finally, the model returns the relation with the maximum $s^\text{avg}_r + b_r$ for single-label prediction, and all relations with higher $s^\text{avg}_r + b_r$ than \textsc{na} for multi-label prediction.
We use classwise kNN because it is more suitable for RE datasets, where the label distribution is usually long-tailed~\cite{zhang-etal-2019-long}.

%% file: experiments.tex
\section{Experiments}

We evaluate our proposed method with a focus on low-resource RE (Sections~\ref{ssec:dataset}-\ref{ssec:results}), and present detailed analyses (Section~\ref{ssec:ablation}) and visualization (Section~~\ref{ssec:visualization}) to justify method design choices.

\subsection{Datasets}\label{ssec:dataset}
We conduct experiments with two document-level RE datasets.
The \textbf{BioRED} dataset~\cite{luo2022biored} is a manually labeled single-label RE dataset in the biomedical domain.
The entity pairs are classified into 9 types (including an \textsc{na} type indicating no relation).
It has a training set consisting of 400 documents, which we use in finetuning.
For pretraining, we use the PubTator Central corpus~\cite{wei2019pubtator}, which annotates the PubMed corpus with entity mentions and their named entity types.
The \textbf{Re-DocRED} dataset~\cite{tan2022revisiting} is a multi-label large-scale dataset of the general domain.
It is a relabeled version of the DocRED dataset~\cite{yao-etal-2019-docred}.
Re-DocRED addresses the incomplete annotation issue of DocRED, where a large percentage of entity pairs are mislabeled as \textsc{na}.
The entity pairs in Re-DocRED are classified into 97 types (incl. \textsc{na}).
It has a training set consisting of 3,053 documents, which we use in finetuning.
For pretraining, we use the distantly labeled training set provided by DocRED, which consists of 101,873 documents.
We remove the relation labels and use our improved MTB to pretrain the model.

\subsection{Experimental Setup}
\stitle{Model configurations.}
We implement our models using Hugging Face Transformers~\cite{wolf-etal-2020-transformers}.
We use AdamW~\cite{loshchilov2018decoupled} in optimization with a weight decay of 0.01.
During pretraining, we use a batch size of 16, a learning rate of 5e-6, a temperature of 0.05, and epochs of 3 and 10 for BioRED and DocRED, respectively.
During finetuning, we use a batch size of 32, a learning rate of 5e-5, and epochs of 100 and 30 for BioRED and DocRED, respectively.
The temperatures in MCCL are set to $\tau_1=\tau_2=0.2$ for BioRED and $\tau_1=0.01, \tau_2=0.03$ for DocRED.
We search $k$ from $\{1, 3, 5, 10, 20\}$ for classwise kNN using the development set\footnote{For low-resource setting with $p$\% of training data, we sample $p$\% of development data as the development set.}.
We run experiments with Nvidia V100 GPUs.

\stitle{Evaluation settings.}
In this work, in addition to the standard full-shot training, we consider low-resource settings.
To create each of the settings, we randomly sample a fixed proportion $p$\% of the entity pairs from the training set as our training data, and use the original test set for evaluation.
We use the same evaluation metrics as the original papers.
We use micro-$F_1$ for BioRED, and micro-$F_1$ and micro-$F_1$-Ign for Re-DocRED.
The micro-$F_1$-Ign removes the relational facts in the test set that have appeared in training.

\stitle{Compared methods.}
We experiment with the following finetuning objectives:
(1) \textbf{Lazy learning}, which directly uses the pretrained embedding and training data to perform kNN without finetuning; (2) \textbf{Cross-entropy loss} (CE), which adds a softmax classifier on top of PLM and uses cross-entropy loss to finetune the model; (3) \textbf{Supervised contrastive loss} (SupCon); and (4) \textbf{Multi-center contrastive loss} (MCCL).
In inference, classwise kNN is used for all methods except for CE.
Note that as SupCon does not apply to multi-label scenarios, we only evaluate it on BioRED.
For each objective, we also evaluate the PLM before and after MTB pretraining.
We use different PLMs as the backbone of the model, namely PubmedBERT$_\textsc{base}$ for BioRED and BERT$_\textsc{base}$ for Re-DocRED, which are pretrained on the biomedical and general domains, respectively.

\begin{table*}[!t]
    \centering
    \scalebox{0.87}{
    \begin{tabular}{ccccccccccc}
    \toprule
     Encoder & Objective& \multicolumn{2}{c}{1\%}& \multicolumn{2}{c}{5\%} & \multicolumn{2}{c}{10\%}& \multicolumn{2}{c}{100\%}\\
    &&$F_1$& $F_1$-Ign& $F_1$& $F_1$-Ign&$F_1$& $F_1$-Ign&$F_1$& $F_1$-Ign\\
    \midrule
    \multirow{4}{*}{PLM}& Lazy& 15.6& 14.9&20.1& 19.4& 21.6& 19.2& 28.7& 28.0\\
    & CE& 40.3& 38.9&54.1& 52.6& 61.3& 60.3& 70.9& 69.4\\
    & MCCL& 44.7& 43.1&59.1& 57.5& 63.2& 61.8& 68.2& 66.7\\
    \midrule
    \multirow{4}{*}{MTB}& Lazy& 35.2& 34.4& 44.7& 43.4& 47.3& 46.2& 54.1& 52.9\\
    & CE& 42.3& 40.7&57.9& 56.4& 62.9& 61.4& \textbf{71.2}& \textbf{69.9}\\
    & MCCL& \textbf{46.4}& \textbf{44.5}&\textbf{59.7}& \textbf{58.2}& \textbf{63.8}& \textbf{62.1}& 69.3& 67.9\\
    \bottomrule
    \end{tabular}}
    \caption{Results on the test set of Re-DocRED.}
    \label{tab:docred_res}
\end{table*}

\begin{table}[!t]
    \centering
    \scalebox{0.87}{
    \begin{tabular}{cccccc}
    \toprule
     Encoder & Objective& 1\%& 5\%& 10\%& 100\%\\
    \midrule
    \multirow{4}{*}{PLM}& Lazy& 14.5& 17.6& 18.8& 28.3\\
    & CE&24.1& 35.4& 42.5& 57.7  \\
    & SupCon& 20.0& 30.9& 38.0& 52.2  \\
    & MCCL& 20.8& 41.3& 45.5& 55.1  \\
    \midrule
    \multirow{4}{*}{MTB}& Lazy& 24.3& 28.4& 34.4& 36.7\\
    & CE& 28.6& 41.2& 49.8& \textbf{61.5}  \\
    & SupCon& 24.4& 29.1& 31.4& 43.1 \\
    & MCCL& \textbf{34.6}& \textbf{48.5}& \textbf{54.2}& 60.8  \\
    \bottomrule
    \end{tabular}}
    \caption{$F_1$ on the test set of BioRED.}
    \label{tab:biored_res}
\end{table}

\subsection{Main Results}\label{ssec:results}
The results on the test sets of Re-DocRED and BioRED are shown in Table~\ref{tab:docred_res} and Table~\ref{tab:biored_res}, respectively.
All results are averaged for five runs of training using different random seeds.
Overall, the combination of MTB and MCCL achieves the best performance in low-resource settings where 1\%, 5\%, and 10\% of relation-labeled data are used.
Further, when using the same MTB-based representations, MCCL shows better results than CE in low-resource settings.
It shows that in low-resource settings, MCCL can better leverage the pretraining knowledge with a well-aligned finetuning objective.
However, this improvement diminishes when abundant labeled data are available, as MCCL underperforms CE on both datasets with full training data on both datasets. In addition, we observe that MTB pretraining consistently improves MCCL and CE on both datasets.
These results demonstrate the effectiveness of MTB pretraining for more precise document-level RE with less needed end-task supervision.

Considering other training objectives, we observe that lazy learning produces meaningful results.
On both datasets, the results of lazy learning based on MTB with 10\% of data are comparable to finetuning with 1\% of data.
This shows that the entity pair embedding pretrained on unlabeled corpora contains knowledge that can be transferred to unseen relations.
We also observe that SupCon using kNN-based inference underperforms both CE and MCCL on BioRED, showing that its one-cluster assumption hurts the knowledge transfer.

\subsection{Ablation Study}
\label{ssec:ablation}


\begin{table}[!t]
    \centering
    \scalebox{0.87}{
    \begin{tabular}{p{3.6cm}ccc}
    \toprule
     Pretraining Objective & 1\%& 10\%& 100\%\\
     \midrule
     PLM& 20.8& 45.5& 55.1\\
     vanilla MTB& 22.9& 45.0& 56.0 \\
     our MTB& 34.6& 54.2& 60.8 \\
     \midrule
     w/o $\mathcal{L}_\text{rel}$& 21.0& 47.1& 56.7 \\
     w/o $\mathcal{L}_\text{self}$& 24.1& 49.3& 58.6\\
     w/o $\mathcal{L}_\text{mlm}$& 32.9& 50.2& 58.2 \\
    \bottomrule
    \end{tabular}}
    \caption{$F_1$ on the test set of BioRED with different pretraining objectives. We use MCCL in finetuning.}
    \label{tab:pretrain}
\end{table}

\stitle{Pretraining objectives.}
We analyze the effectiveness of our proposed pretraining losses in Section~\ref{ssec:pretraining}.
To do so, we pretrain the model with one loss removed at a time while keeping the finetuning setup on BioRED fixed with the MCCL.
The results are shown in Table~\ref{tab:pretrain}.
Overall, we observe that all losses are effective.
If we remove all proposed techniques and use the vanilla MTB pretraining objective of binary pairwise classification, the results are only slightly better or even worse.
Among the techniques, removing $\mathcal{L}_\text{rel}$ leads to the largest performance drop, showing that MTB-based pretraining is critical to improve low-resource RE.
Removing $\mathcal{L}_\text{self}$ also leads to a large performance drop.
It is because $\mathcal{L}_\text{self}$ encourages the model to learn more discriminative features that lead to less collapsed representations.
Our finding aligns with recent studies in computer vision~\cite{islam2021broad,chen2022perfectly}, showing that reducing collapsed representations with self-supervised contrastive learning improves the transferability to downstream tasks.


\begin{figure}[!t]
    \centering
    \scalebox{0.27}{
    \includegraphics{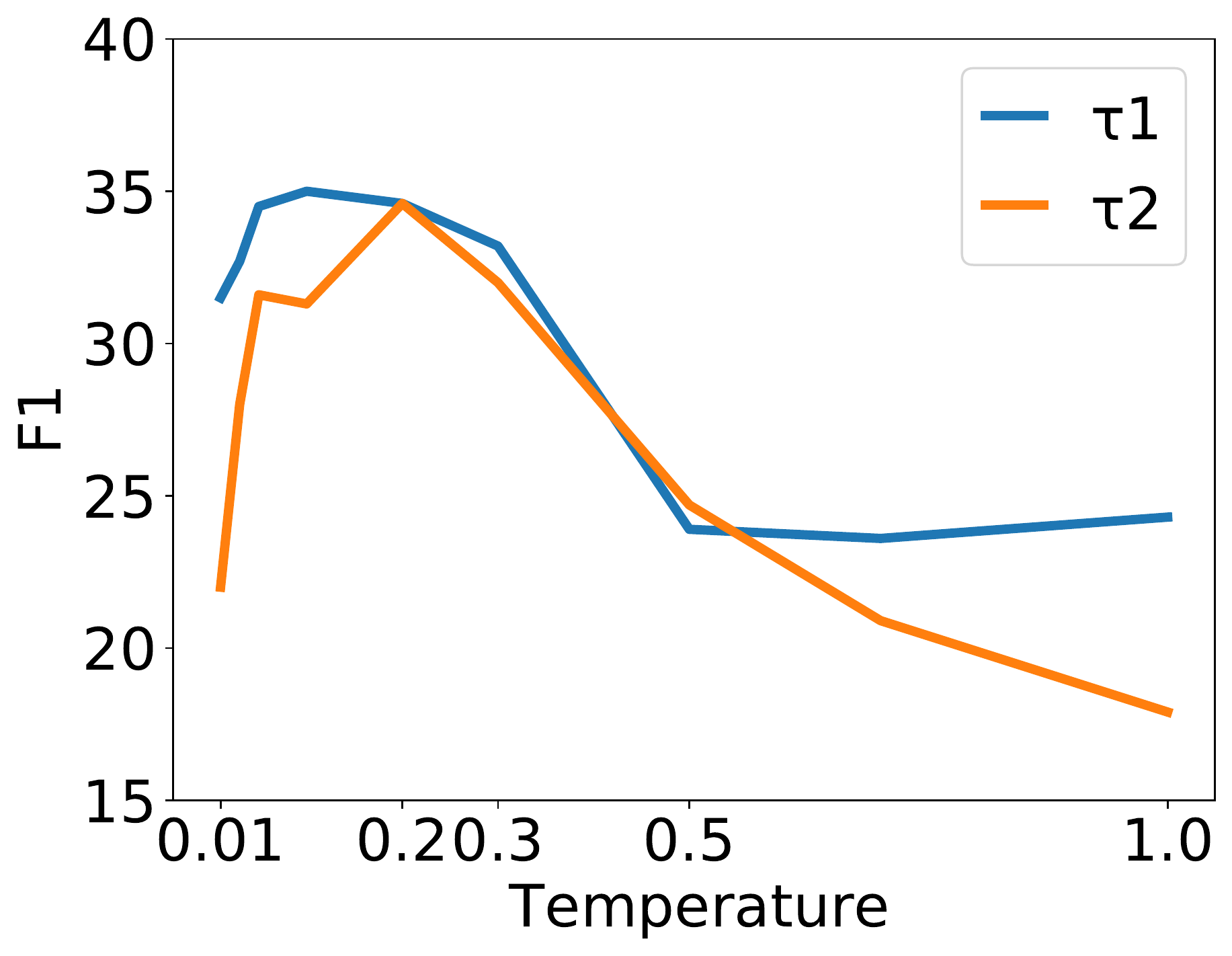}}
    \caption{$F_1$ using different temperatures on 1\% of BioRED.}
    \label{fig:temperature}
    \vspace{-1em}
\end{figure}

\stitle{Performance w.r.t. different temperatures.}
We discuss the impact of two temperatures in MCCL.
In MCCL, $\tau_1$ controls the weighting of instances.
With a very small $\tau_1$, each instance will only form a cluster with its nearest neighbor in the batch, while with very large $\tau_1$, instances of the same relation will collapse to the same cluster.
$\tau_2$ controls the importance of hard instances,
which is also used in other contrastive losses (e.g., $\tau$ in Eq.~\eqref{eq:contra}).
\citet{wang2021understanding} observe that small $\tau_2$ makes the model focus more on hard instances, while \citet{khosla2020supervised} observe that too small $\tau_2$ leads to numerical instability.
We show the results of using different temperatures in Figure~\ref{fig:temperature}, where we keep one temperature fixed and change the other.
For $\tau_1$, we find that using large temperature harms the performance, showing that our multi-cluster assumption improves low-resource RE.
For $\tau_2$, we observe that both small and large values impair the performance, which is aligned with prior observations.

\begin{figure}[!t]
    \centering
    \scalebox{0.27}{
    \includegraphics{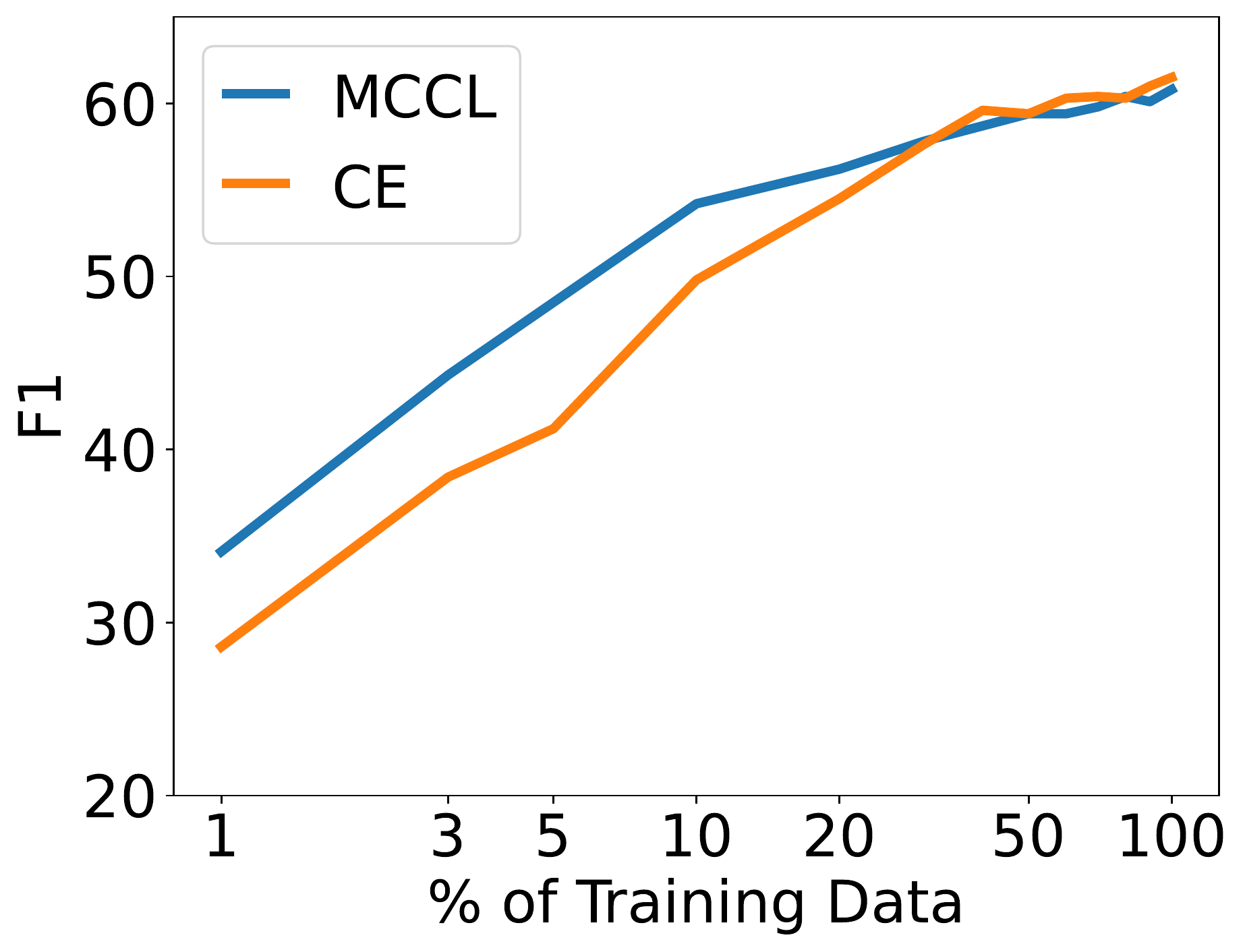}}
    \caption{$F_1$ under different percentages of BioRED training data.}
    \label{fig:percentage}
\end{figure}

\stitle{Performance w.r.t. different amount of data.}
The main results show that MCCL outperforms CE in the low-resource setting, while slightly underperforming CE when full training data is used.
We further evaluate MCCL and CE using different amounts of end-task data.
We experiment on BioRED and use the entity pair embedding pretrained with MTB.
Results are shown in Figure~\ref{fig:percentage}.
We observe that MCCL consistently outperforms CE by a large margin when 
less than 20\% of training data is used, while it performs similarly or worse than CE after that.
It again demonstrates the effectiveness of MCCL in low-resource RE.
However, as the pretraining and finetuning are based on different tasks, fully adapting the model to downstream data by CE results in similar or better performance in data-sufficient scenarios.

\subsection{Visualization}
\label{ssec:visualization}

\begin{figure}
    \centering
    \scalebox{0.3}{
    \includegraphics{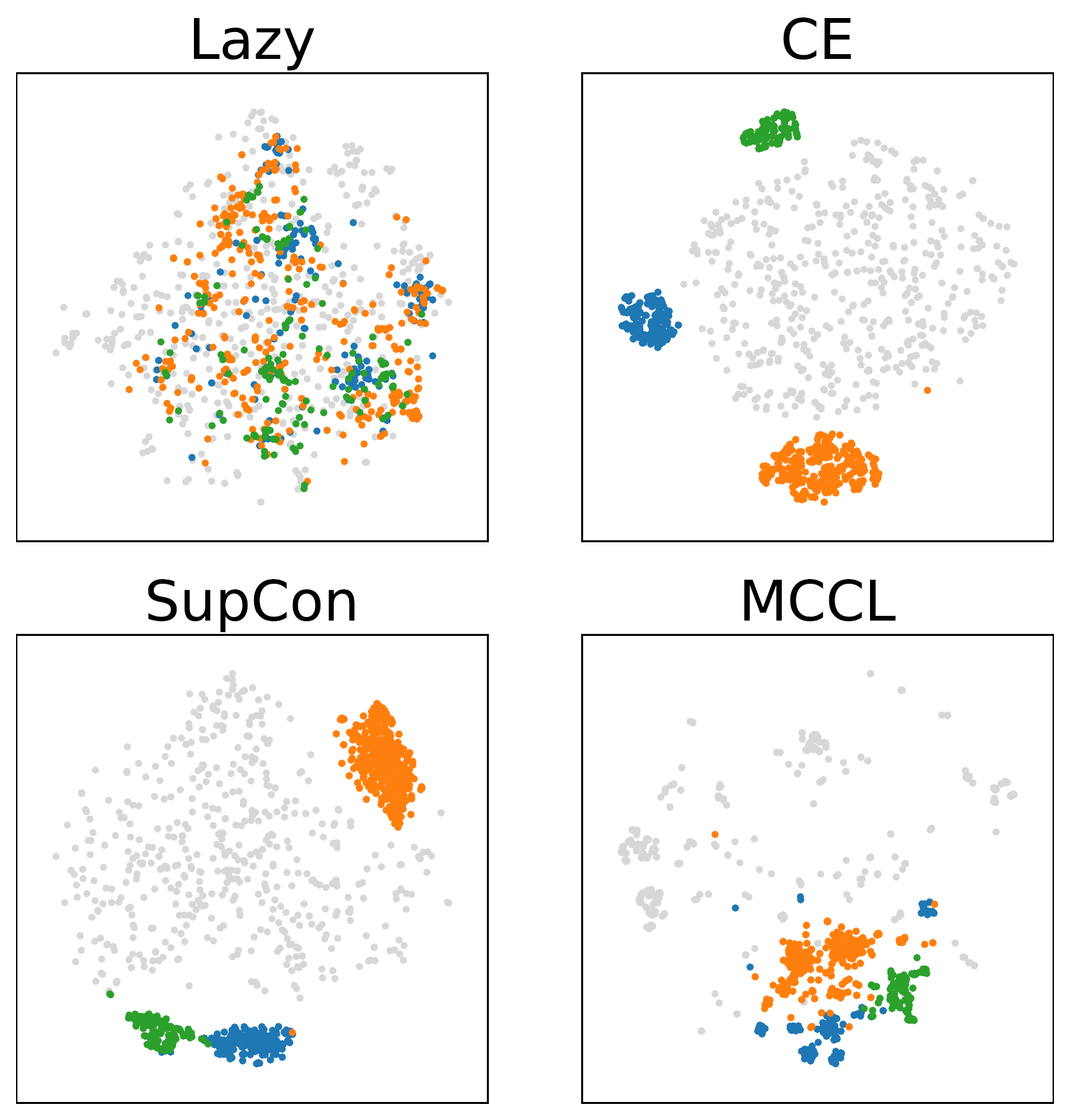}}
    \caption{Visualization of entity pair embedding finetuned with different objectives on BioRED. \textsc{na} instances are shown in grey.}
    \label{fig:vis}
    \vspace{-1em}
\end{figure}

Figure~\ref{fig:vis} shows the t-SNE~\cite{van2008visualizing} projection of entity pair embedding finetuned with different objectives on BioRED. 
For clarity, we visualize the embedding of the four most frequent relations in BioRED with different colors, including the \textsc{na} class shown in grey.
The visualization shows that both CE and SupCon learn one cluster for each relation, while lazy learning and MCCL, as expected, generate multiple small clusters for a relation.
This observation indicates that MCCL can better align with the pretraining objective, further explaining its better performance in low-resource settings.

%% file: conclusion.tex
\section{Conclusion}
In this paper, we study self-supervised learning for document-level RE.
Our method conducts an improved MTB pretraining objective that acquires cheap supervision signals from large corpora without relation labels.
To bridge the gap between pretraining and end-task finetuning, we propose a continual contrastive finetuning objective, in contrast to prior studies that typically use classification-based finetuning, and use kNN-based inference.
As pretrained representation may form multi-cluster representation, 
we further propose a multi-center contrastive loss 
that aligns well with the nature of the pretrained representation.
Extensive experiments on two document-level RE datasets demonstrate the effectiveness of these key techniques in our method.
Future work is adapting our method to other tasks in information extraction, such as n-ary relation extraction, named entity recognition, typing, and linking.

\section*{Limitations}
The main limitation of MCCL is the requirement of a sufficiently large batch size in training (32 documents in our experiments), leading to a need for large GPU memory.
This is because MCCL uses in-batch entity pairs for contrastive learning, and a small batch size does not provide enough instances to form multiple clusters.
In addition, we need to store the entity pair embedding of the whole training set for kNN-based inference, which is less memory-efficient than CE.

%% file: appendix.tex
\appendix
\begin{center}
    {
    \Large\textbf{Appendices}
    }
\end{center}
\section{Data Preparation}
\label{ssec:data_preparation}
We acquire positive and negative pairs from the document corpus.
We regard two entity pairs $(e_{s_1}, e_{o_1}), (e_{s_2}, e_{o_2})$ in different documents as a positive pair if they share the same subject and object entities, respectively (i.e., $e_{s_1}=e_{s_2}, e_{o_1}=e_{o_2}$), and otherwise negative.

However, for a large corpus, the number of such positive pairs is enormous.
For instance, in biomedical RE pretraining, we extract 37 billion positive pairs in total.
Using all these pairs in pretraining is computationally infeasible.
Therefore, we select positive pairs as follows.
Denote the number of documents mentioning an entity $e$ or an entity pair $(e_s, e_o)$ as $N(e)$ and $N(e_s, e_o)$, respectively, we use two metrics, $\text{frequency} = N(e_s, e_o)$ and $\text{PMI} = \frac{N(e_s, e_o)}{N(e_s) \times N(e_o)}$, to measure the popularity of entity pairs
The frequency measures how often $e_s$ and $e_o$ co-occur.
The PMI measures whether $e_s$ and $e_o$ have a strong association.
In pretraining, we first discard the entity pairs with frequency $< N_\text{threshold}$, and then use the positive pairs constituted by the top $K$ entity pairs measured by their PMIs.
We set the frequency threshold to be 16 and 3 for BioRED and DocRED, respectively, and use the top 5,000 entity pairs in pretraining.

Besides, as MTB is fully self-supervised, the information of whether two relations mentions correspond to the same relation type is not available, but it is assumed that at least entity pairs with different subject or object types are likely to be of different relation types and can therefore be used as negative pairs. Such use of entity types to filter the pairs has indeed been shown a strong feature for RE~\cite{zhong-chen-2021-frustratingly,zhou2021improved}.
We only use two entity pairs with different subject or object entity types as negatives.
While the entity type based filtering may also discard some hard negatives, our experiment (see Section~\ref{ssec:more_experiments}) shows improved results, meaning that its benefits outweigh the disadvantages.

\section{Adaptation to Multi-label RE}
\label{ssec:multilabel_RE}
It is noteworthy that in some RE tasks, such as DocRED, one entity pair may have multiple relation labels, in which case the cross-entropy loss does not apply.
Therefore, for multi-label scenarios, we substitute cross-entropy loss (also the softmax in MCCL) with the adaptive thresholding loss proposed by~\citet{zhou2021document}.
Specifically, denote the logits as $l$ (the input to softmax in cross-entropy loss), the set of positive relations as $\mathcal{P}$ (except \textsc{na}), and the set of the remaining relations except for \textsc{na} as $\mathcal{N}$, the adaptive thresholding loss is formulated as:
\begin{align*}
    \mathcal{L}_1 &= -\sum_{r\in \mathcal{P}} \log \left( \frac{e^{l_r}}{\sum_{r' \in \mathcal{P} \cup \{\textsc{na}\}} e^{l_{r'}}} \right), \\
    \mathcal{L}_2 &= - \log \left( \frac{e^{l_\textsc{na}}}{\sum_{r' \in \mathcal{N} \cup \{\textsc{na}\}} e^{l_{r'}}} \right), \\
    \mathcal{L}_\text{at} &= \mathcal{L}_1 + \mathcal{L}_2.
\end{align*}
This loss encourages the logits of positive relations to be higher than \textsc{na}, and the logits of other relations to be lower than \textsc{na}.
In prediction, the model returns the relations with higher logits than \textsc{na} as predictions, or return \textsc{na} if none of such relations exist.

\section{More Experiments}
\label{ssec:more_experiments}
\begin{figure}[!t]
    \centering
    \scalebox{0.27}{
    \includegraphics{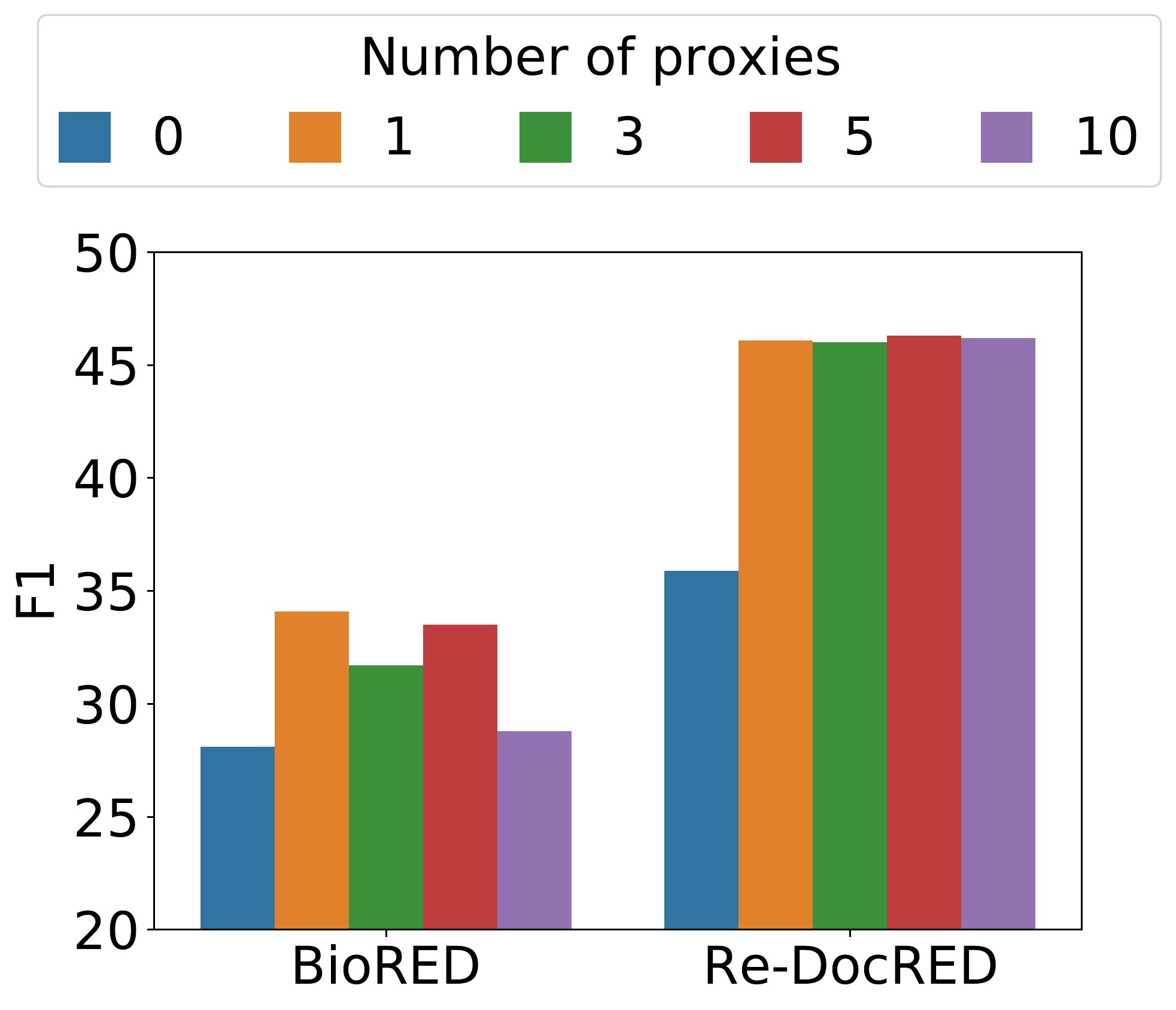}}
    \caption{$F_1$ achieved with different numbers of proxies on 1\% BioRED.}
    \label{fig:proxy}
\end{figure}

\stitle{Performance w.r.t. number of proxies.}
We evaluate MCCL with different number of proxies.
When no proxy is used, we ignore the relations that do not appear in the current batch.
The $F_1$ on both BioRED and Re-DocRED in the 1\% low-resource setting is shown in Figure~\ref{fig:proxy},
indicating that adding proxies improves $F_1$ significantly on both datasets.
Using one proxy for each relation achieves an increase of 6.0\% in $F_1$ on BioRED, and a larger increase of 10.2\% in $F_1$ on Re-DocRED.
Such a difference of increment is due to the fact that Re-DocRED is more long-tailed, where 97\% of instances are \textsc{na} compared to 80\% in BioRED.
We also observe that adding more proxies achieves similar or even worse results.
These results make sense as the proxies are mainly in the place of long-tail relations that do not appear in the batch, and these relations contain too few instances to form multiple clusters.

\stitle{Coarse-to-fine evaluation.}
To give another illustration of showing that MCCL learns multiple clusters,
we experiment with it on 1\% of BioRED in a coarse-to-fine setting.
Specifically, we merge all relations except \textsc{na} into one relation in finetuning, and apply kNN inference using the original labels.
We find that MCCL achieves an F1 of 30.3\%, which is even better than CE with all relations provided. However, if we remove the instance weights in MCCL to degrade it to one-cluster, the $F_1$ constantly degrades in finetuning.
It shows that multi-cluster assumption helps preserve the fine-grained relation information in pretrained representation.

\stitle{Other ablation studies.}
We analyze the effectiveness of entity type filtering in Section~\ref{ssec:data_preparation}.
Results are shown in Table~\ref{tab:pretrain_app}.
Removing entity type filtering degrades performance significantly.
It shows that entity type filtering can remove a lot of false negatives in pretraining and greatly improves the pretrained model.

Besides, as the main results have demonstrated the effectiveness of MCCL in finetuning, we wonder whether MCCL can also lead to improved pretraining.
To do so, we replace the InfoNCE loss in Eq.~\eqref{eq:contra} by MCCL and regard different entity pairs as different classes.
The results are comparable or slightly worse in contrast to using $\mathcal{L}_\text{rel}$, showing that the multi-cluster assumption of MCCL does not necessarily help pretraining.

\begin{table}[!t]
    \centering
    \scalebox{0.87}{
    \begin{tabular}{p{3.6cm}ccc}
    \toprule
     Pretraining Objective & 1\%& 10\%& 100\%\\
     \midrule
     PLM& 20.8& 45.5& 55.1\\
     vanilla MTB& 22.9& 45.0& 56.0 \\
     our MTB& 34.6& 54.2& 60.8 \\
     \midrule
     w/o entity type filtering& 25.1& 48.6& 58.1 \\
     Replace $\mathcal{L}_\text{rel}$ by $\mathcal{L}_\text{mccl}$& 34.7& 52.5& 58.8\\
    \bottomrule
    \end{tabular}}
    \caption{$F_1$ on the test set of BioRED.}
    \label{tab:pretrain_app}
\end{table}